\documentclass[runningheads]{llncs}
\usepackage{graphicx}

\usepackage{tikz}
\usepackage{comment}
\usepackage{amsmath,amssymb} 
\usepackage{color}
\usepackage{epsfig}
\usepackage[export]{adjustbox}
\usepackage{multirow}
\usepackage{diagbox}
\usepackage{cite}

\usepackage[accsupp]{axessibility}  

\usepackage[breaklinks=true,bookmarks=false,colorlinks,]{hyperref}

\begin{document}
\pagestyle{headings}
\mainmatter
\def\ECCVSubNumber{6977}  

\title{ViDaS: Video Depth-aware Saliency Network} 

%
\author{Ioanna Diamanti \and
Antigoni Tsiami \and
Petros Koutras \and
Petros Maragos}
\authorrunning{I. Diamanti et al.}
%
\institute{School of E.C.E., National Technical University of Athens, Greece \\
\email{i.diamanti@outlook.com}\\
\email{\{antsiami, pkoutras, maragos\}@cs.ntua.gr}}
\maketitle

\begin{abstract}
We introduce ViDaS, a two-stream, fully convolutional Video, Depth-Aware Saliency network to address the problem of attention modeling ``in-the-wild", via saliency prediction in videos. Contrary to existing visual saliency approaches using only RGB frames as input, our network employs also depth as an additional modality.
The network consists of two visual streams, one for the RGB frames, and one for the depth frames. Both streams follow an encoder-decoder approach and are fused to obtain a final saliency map. The network is trained end-to-end and is evaluated in a variety of different databases with eye-tracking data, containing a wide range of video content. 
Although the publicly available datasets do not contain depth, we estimate it using three different state-of-the-art methods, to enable comparisons and a deeper insight.
Our method outperforms in most cases state-of-the-art models and our RGB-only variant, which indicates that depth can be beneficial to accurately estimating saliency in videos displayed on a 2D screen. 
Depth has been widely used to assist salient object detection problems, where it has been proven to be very beneficial. Our problem though differs significantly from salient object detection, since it is not restricted to specific salient objects, but predicts human attention in a more general aspect. These two problems not only have different objectives, but also different ground truth data and evaluation metrics. To our best knowledge, this is the first  competitive deep learning video saliency estimation approach that combines both RGB and Depth features to address the general problem of saliency estimation ``in-the-wild". The code will be publicly released.
\end{abstract}

\section{Introduction}


Video saliency detection is the task of estimating human eye fixations when perceiving dynamic scenes. The problem of modeling human attention has gained more and more interest over the recent years due to its important contribution in a variety of applications such as video summarization, video compression, virtual reality and robotics. In parallel, the development of Deep Learning techniques and especially Convolutional Neural Networks (CNNs) has helped achieve remarkable results in various computer vision problems such as image segmentation, classification, saliency estimation etc.

\begin{figure}[t]
\begin{center}
   \includegraphics[width=0.5\linewidth]{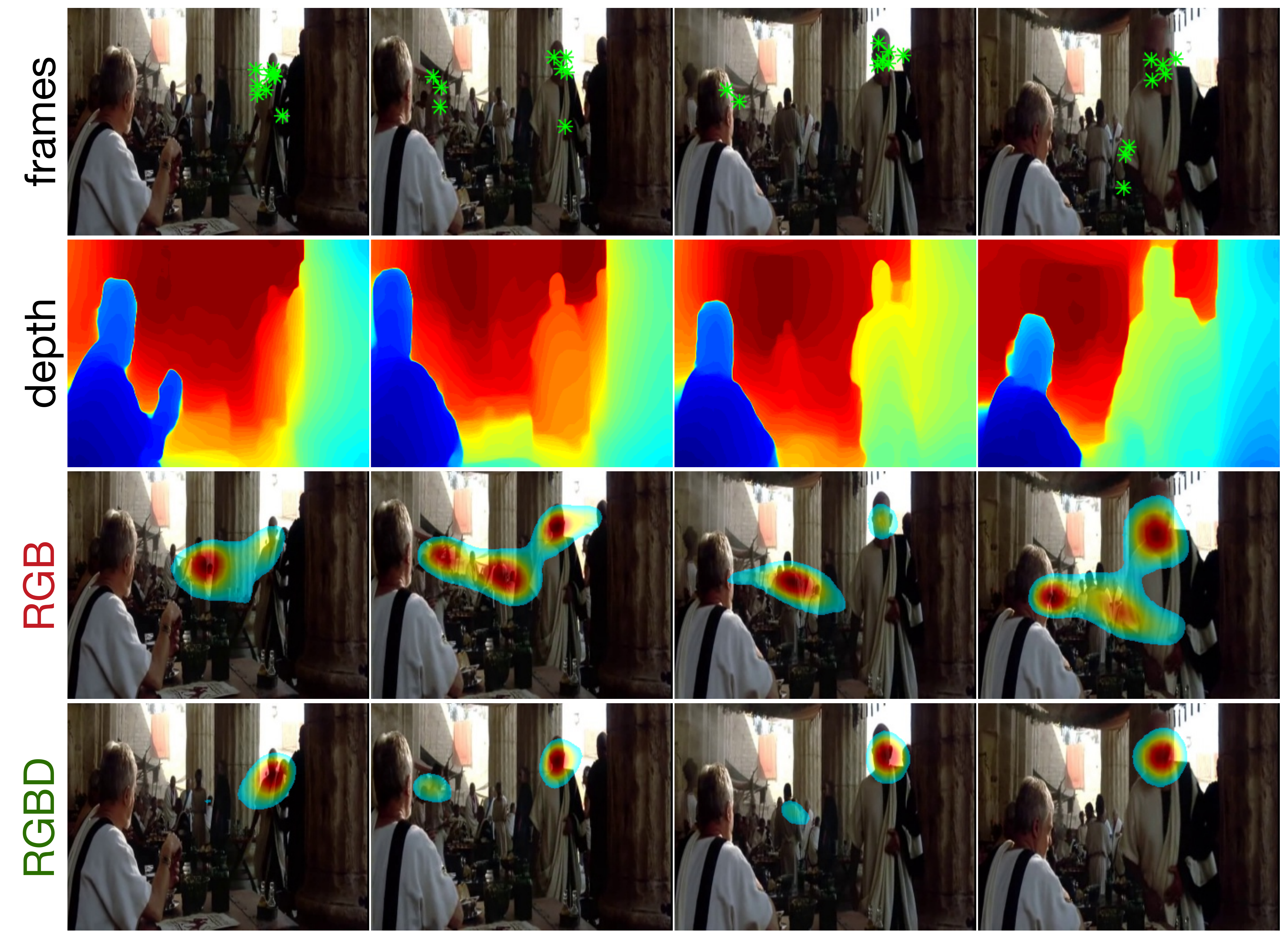}
\end{center}
\vspace{-0.2cm}
   \caption{Frames with their eye-tracking data from a Hollywood movie, along with the frames' estimated depth. The third row depicts saliency maps produced by RGB-only saliency network, while the last row is the output of our proposed ViDaS RGB-D network, which succeeds in better predicting human attention.}
\label{fig:figure1}
\vspace{-0.7cm}
\end{figure}

Video saliency estimation is considered to be a more challenging task compared to static image saliency prediction. That is because in videos we need to accurately extract both spatial and temporal features and fuse them effectively in order to obtain a final saliency map. Previously introduced methods widely depend on LSTMs and Convolutional Neural Networks \cite{salsac} in order to compute the spatio-temporal features and are mostly focusing on the visual information provided by the various eye-tracking datasets.

Human attention and perception is however deeply affected by many different cues that can be present in a video scene and awake various human senses. Recent research and studies \cite{leifman2017learning, Piao_2019_ICCV, Tsiami_2020_CVPR, tavakoli2019dave} indicate that such cues can be the depth information as well as the audio. As illustrated by Fig.\ref{fig:figure1} incorporating such information can assist in locating salient regions. 

Depth information is inseparably connected to visual stimuli since the human brain has the ability to detect the different objects present in a video scene or image and estimate their relative distance. This cue is naturally perceived and processed by the brain and this led us to the idea of integrating depth and RGB information in a single network to assess whether it could improve saliency prediction in videos. Moreover, nowdays depth can be easily captured either by depth sensors, that have started appearing in many everyday life devices such as mobile phones, or by employing modern deep learning methods. Fig.~\ref{fig:figure1} highlights and motivates the problem we investigate. A person is coming towards another person in a movies scene. This information is clearly captured by depth, and is mirrored in the results, where RGB alone could not predict attention as well as the combination of RGB and Depth does.

We propose ViDaS, a video depth-aware saliency network that efficiently combines RGB and Depth information in order to predict saliency in video scenes. Our approach extends and improves the RGB variant of an existing state-of-the-art video saliency network proposed in~\cite{Tsiami_2020_CVPR}. We include a second stream that takes as an input the produced depth maps of the corresponding RGB frames given as input to the first stream and extracts multi-scale features. The features obtained from both streams, are fused effectively together in order to obtain a final saliency map. Our problem differs significantly from salient object detection, since it is not restricted to specific salient objects, but predicts human attention in a more general aspect. These two problems not only have different objectives, but also different ground truth data and evaluation metrics. 

To the best of our knowledge, there is no eye-tracking dataset containing depth information for video sequences except for~\cite{leifman2017learning}, where a limited data collection using a Kinect camera took place, but it is not publicly available. Therefore, we use a robust state-of-the-art  depth estimation network \cite{DBLP:journals/corr/abs-1907-01341} capable of accurately predicting both indoor and outdoor scenes, in order to extract depth from the 2D RGB frames of the eye-tracking video databases. 
We investigate 3 different methods for depth extraction in order to assess depth contribution in the network's performance. Experimentation is carried out in 9 different databases with a large variety of video content, including sports, movies, user-made videos, documentaries, meeting scenes, etc. For comparison purposes, we employ different training setups, including our RGB-only variant, and we compare our performance with 11 different state-of-the-art models.  Results indicate that depth successfully contributes to saliency modeling and is a useful modality to employ for attention modeling. ViDaS RGB-D performance accross the various databases and unseen datasets indicates that our model is capable of modeling saliency ``in-the-wild". 

\section{Related Work}

\subsection{RGB Saliency}

Early CNN video saliency estimation approaches have been based on the adaptation of pretrained CNN models initially proposed for visual action recognition tasks \cite{Kummerer2014b,vig2014large}. Later, in~\cite{Pan_2016_CVPR} shallow and deep CNNs were trained for saliency prediction while in~\cite{huang2015salicon, jetley2016end} training was performed by optimizing common saliency evaluation metrics. Long-Short Term Memory (LSTMs) and Generative Adversarial Networks (GANs) have also been developed for saliency prediction, e.g. LSTMs for spatial-only saliency in static images~\cite{cornia2018predicting} and spatio-temporal in videos~\cite{wang2018revisiting,wang2019revisiting,Linardos2019}, as well as GANs in~\cite{Pan_2017_SalGAN}. Multi-level saliency information from different layers through skip connections has been employed in~\cite{wang2018deep}.
More recently, the TASED method~\cite{Min_2019_ICCV} employs a 3D fully-convolutional network with temporal aggregation, based on the assumption that the saliency map of any frame can be predicted by considering a limited number of past frames. Also, the authors of~\cite{unisal} essentially unify spatial and spatio-temporal saliency, i.e. image and video saliency into a joint saliency network by introducing four novel domain adaptation techniques.
In order to improve saliency estimation in videos, some approaches have employed multiple modalities by combining them in multi-stream networks. For example, RGB/Optical Flow (OF) have been both employed in~\cite{bak2017spatio} and more recently in~\cite{lai2019video}, RGB/Audio in~\cite{Tsiami_2020_CVPR, tavakoli2019dave}. Another multi-stream example is multiple subnets, such as objectness/motion \cite{jiang2018deepvs} or saliency/gaze \cite{Gorji_2018_CVPR} pathways.

\subsection{RGB-D Saliency}

Depth has been employed in a variety of computer vision related problems. However, depth-aware saliency estimation in videos in the context of general attention modeling has not been explored as much as in the specific context of salient object detection (SOD). According to a recent survey~\cite{zhou2021rgb}, more than 100 models have used depth along with RGB frames (RGB-D) for SOD, starting back in 2012~\cite{lang2012depth,desingh2013depth} and continuing till today with deep learning models~\cite{zhu2019pdnet, Piao_2019_ICCV, chen2021rgb}. However, SOD tasks concentrate solely on finding salient objects in a video scene. On the other hand, saliency estimation for attention modeling is a different problem, because it focuses on modeling human attention in a video scene by predicting fixation points. Attention might be captured by objects, but it is not limited to well-defined structures. For example, it might be captured by more abstract visual cues, color difference or salient regions within an object. A few past works only have incorporated depth into a visual saliency model~\cite{wang2013computational}. Especially concerning deep learning methods, there are even fewer~\cite{leifman2017learning}. In this work~\cite{leifman2017learning}, RGB, Depth and Optical Flow are used to produce saliency maps employing generative CNNs. Here the architecture is much more naive, with the depth information integrated in the training process by simply including the depthmaps in the same stream as the RGB frames and processing them together in the various spatio-temporal scales. This method was trained and evaluated using a rather small and restricted RGBD video dataset consisting of only 54 videos. Each of these videos necessarily contains multiple levels of depth, which does not allow investigating the behavior and accuracy of the method in the wild, where depth levels in a scene could possibly be fewer. Also this method incorporates the optical flow in the training process, which not only adds computational cost, but also fails to investigate the possible benefit of using depth alone. Talking about depth, existing eye-tracking databases do not contain depth images. However, several methods exist (lightweight networks, depth cameras integrated even in mobile phones, disparity based methods in videos) that enable robust depth extraction from RGB frames at a low computational cost~\cite{DBLP:journals/corr/abs-1907-01341,li2018megadepth, laina2016deeper, hao2018detail}.  In~\cite{DBLP:journals/corr/abs-1907-01341}, a robust depth extraction method has been developed by training and testing on different large datasets (zero-shot cross-training). 

\section{Proposed Method}

The proposed method follows a 3D fully-convolutional encoder-decoder architecture. It consists of two identical spatio-temporal visual streams (encoders), that compute RGB and Depth saliency features, two fully-convolutional spatial decoder modules that perform an effective fusion of these multi-scale features and the appropriate loss function. The method is explained in detail in the following sections.

\subsection{Encoder}
\begin{figure}[t]
\begin{center}
   \includegraphics[width=\linewidth]{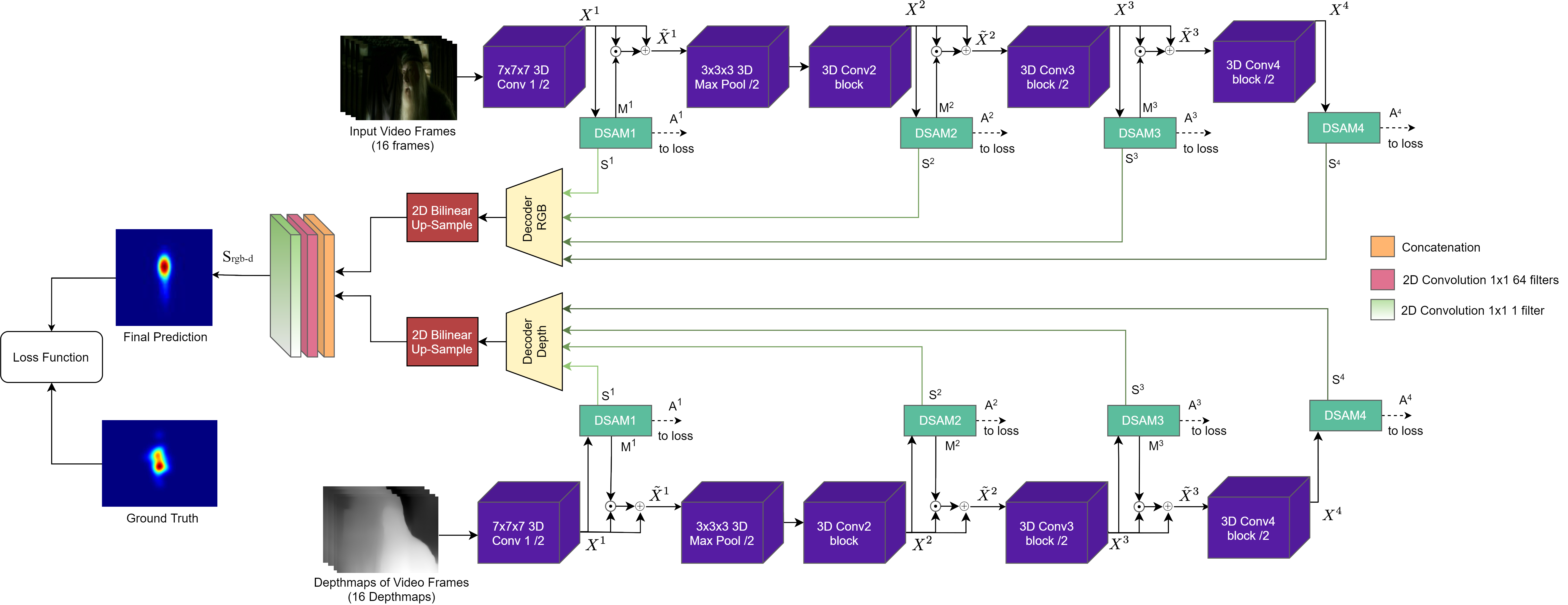}
\end{center}
\vspace{-0.2cm}
   \caption{ViDaS architecture. The network consists of two identical streams, computing RGB and Depth saliency features respectively. The output saliency feature maps from the different scales for RGB and Depth each pass through a Decoder and are fused in the last network layer in order to produce a single saliency map.}
\label{fig:figure2}
\vspace{-0.5cm}
\end{figure}

\begin{figure}[t]
\begin{center}
   \includegraphics[width=0.5\linewidth, frame]{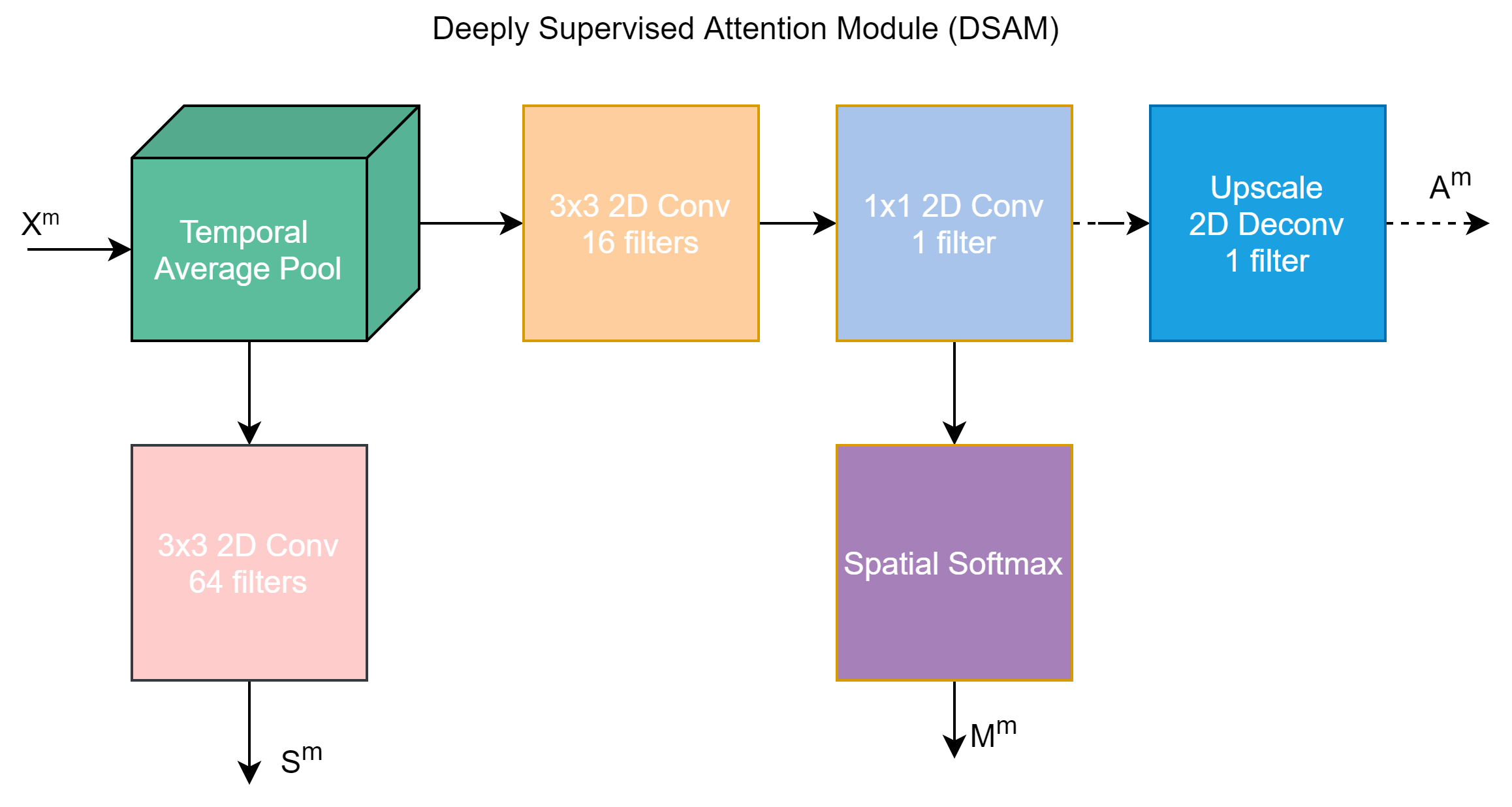}
\end{center}
   \caption{Deeply Supervised Attention Module (DSAM) enhances
            the global network’s representations and provides the multi-level
            saliency maps for spatio-temporal saliency.}
\label{fig:figure3}
\vspace{-0.5cm}
\end{figure}

First, we present the backbone network of our model, displayed in Fig. \ref{fig:figure2} that is used in order to extract the multi-scale spatio-temporal features from both the RGB frames and the Depth. The architecture of the Encoder extends the 3D version of ResNet50, initially proposed for action classification. It consists of 4 3-D fully convolutional blocks that calculate spatio-temporal features in different scales of the input frames represented by $X^{1}, X^{2}, X^{3}, X^{4}$. Each output $X^{m}$ of the convolutional blocks is first refined using an attention mechanism before it continues to the next block. For that purpose we are using the Deeply Supervised Attention Module (DSAM) depicted in Fig. \ref{fig:figure3}

Deep supervision has been formerly used in edge detection\cite{xie2015holistically}, object segmentation \cite{Cae+17} and static saliency \cite{wang2018deep}. The role of DSAM in our model is triple: It is used for enhancing spatial feature representations, for providing the multi-level activation maps $A^{m}$ that will be used to calculate the loss, and finally for providing the multi-level, 64-channel saliency feature maps $S^{m}$ that will be used as an input to the Decoders in order to later obtain a final saliency map. Thus, DSAM parameters $\mathbf{W}^{m}_{am}$ are trained along with all the other trainable parameters of the network.

Figure \ref{fig:figure3} displays the DSAM module architecture at level m. It includes an average pooling in the temporal dimension in order to obtain a 2D representation of the feature maps. The output of the temporal average pooling is then directed to two different paths inside the module. The one path consists of just one spatial convolutional layer that provides the 64-feature saliency maps $S^{m}$. The other path consists of two convolutional layers that finally calculate a single activation map $A^{m}$. A spatial softmax operation applied at the activation map $A^{m}$ yields the attention map $M^{m}$:
\begin{equation}
    M^{m}(x,y) = \frac{\exp(A^{m}(x,y))}{\sum_{x}\sum_{y}\exp(A^{m}(x,y))}
\end{equation}
Finally the activation map $A^{m}$ is up-sampled to the initial dimensions of the input frames using a transposed convolutional layer. The output $X^{m}$ of the corresponding m-level convolutional block of the visual stream is then element-wise multiplied with the attention map $M^{m}$ and added to its initial value in order to enhance its most salient regions, providing the input for the next convolutional block $\tilde{X}^{m}$:
\begin{equation}
    \tilde{X}^{m} = (1 + M^{m}) \odot X^{m},  m = 1,...,4
\end{equation}
where $\odot$ denotes the element-wise multiplication.

\subsection{Decoder}
\begin{figure}[t]
\begin{center}
   \includegraphics[width=0.45\linewidth, frame]{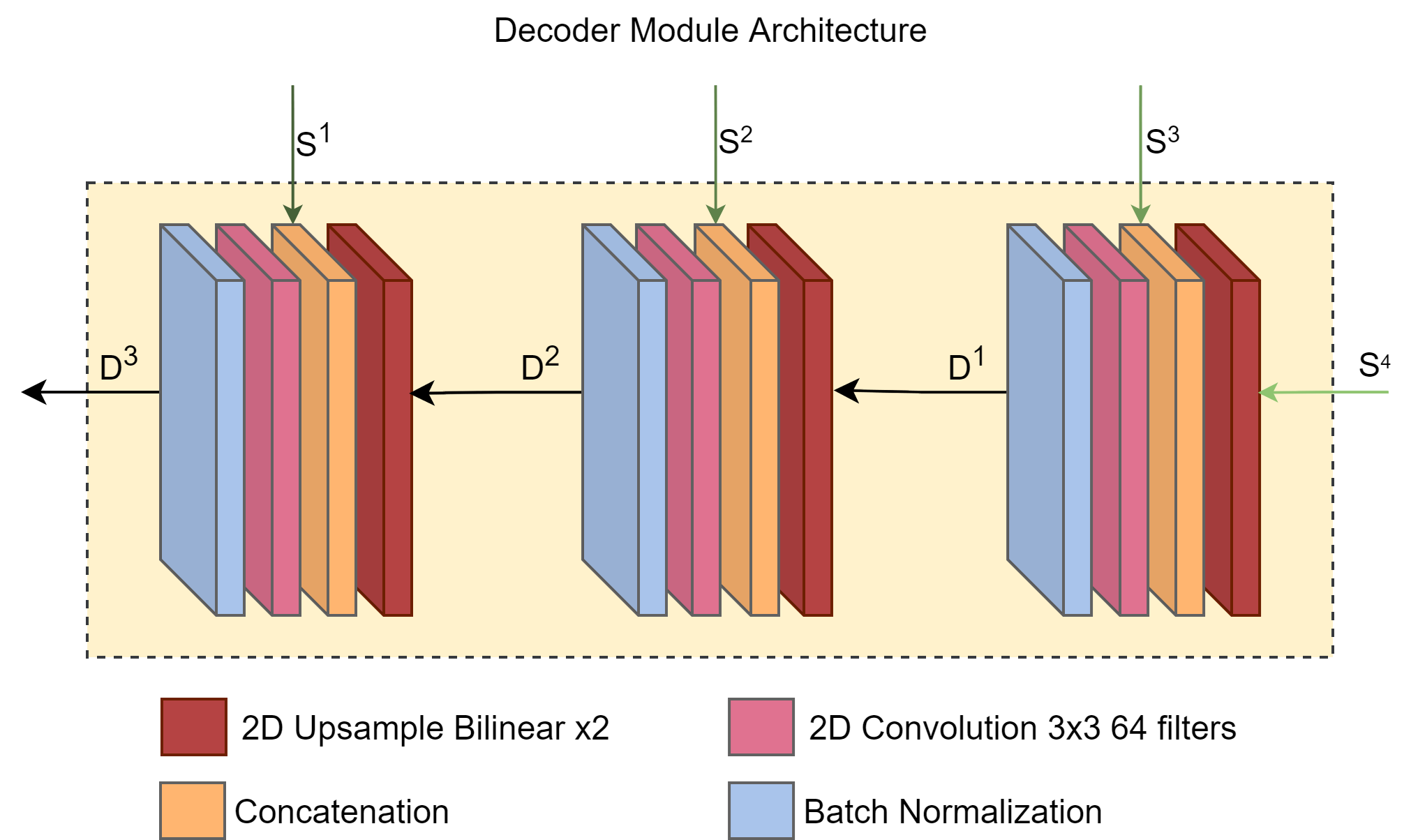}
\end{center}
   \caption{Architecture of the Decoder module. The Decoder is used as the prediction part of the network, which fuses the multi-scale spatio-temporal features to obtain the final saliency map.}
\label{fig:figure4}
\vspace{-0.5cm}
\end{figure}

For each stream, the outputs of its 4 DSAM modules are passed as an input to the Decoder in order to obtain the final saliency map. The Decoder implements a U-Net-like architecture, gradually fusing smaller scale features produced deeper in the network with features calculated at earlier layers. The architecture is illustrated in Fig. \ref{fig:figure4}. The Decoder module consists of three 2D fully-convolutional blocks that are used in order to effectively fuse the multi-scale features produced by the backbone network. The first convolutional block of the Decoder takes as an input the outputs $S^{3}, S^{4}$ of the two last DSAM modules of the encoder. After that, each convolutional block at level $l$ takes as input the output $S^{m}$ of the corresponding DSAM module and the result of the previous block $D^{l-1}$ of the Decoder. To produce the final result, the blocks first contain a 2D bilinear up-sampling layer $\mathrm {U}$ in order to match the input's spatial dimensions. The two inputs $S^{m}, D^{l-1}$ are then concatenated and fused together using a 2D convolutional layer $\mathrm {C}_{l}$, followed by a Batch Normalization layer $\mathrm{BN}_{l}$ to avoid the problem of exploding gradients. No activation function is applied. Thus, the outputs of the convolutional blocks of the Decoder can be estimated by:
\vspace{-0.2cm}
\begin{eqnarray}
    D^{l} = \mathrm {BN}_{l}(\mathrm {C}_{l}(S^{3}, \mathrm {U}(S^{4}))), l = 1 \\
    D^{l} = \mathrm {BN}_{l}(\mathrm {C}_{l}(S^{m}, \mathrm {U}(D^{l-1}))), l = 2,3
\end{eqnarray}
 
\subsection{RGB-D Fusion}
The overall proposed architecture consists of both RGB and Depth streams. The two streams do not interfere with each other until the output $D_{rgb}, D_{d}$ of each Decoder is calculated. That way, each stream can be trained end-to-end concentrated on its separate task which is to learn the appropriate representations from the provided input. By applying a last-layer fusion of RGB and Depth features, we observed that the network could learn when Depth features are beneficial per input case, compared to other approaches we followed. That being said, the fusion between RGB and depth features is done after the decoding is finished. The two outputs $D_{rgb}, D_{b}$ of the two Decoders are up-sampled using bilinear interpolation layer $\mathrm{U}$ to match the input's dimensions. Finally the up-sampled feature maps are fused using 2D convolutional layers to produce the final saliency map $S_{rgb-d}$ of the network:
\begin{equation}
    S_{rgb-d} = \mathrm {F}(\mathrm {U}(D_{rgb}), \mathrm {U}(D_{d}))
\end{equation}
where $\mathrm{F}(\cdot)$ denotes the concatenation and convolutional fusion of the Decoder outputs.

\vspace{-0.2cm}
\subsection{Saliency Loss}

For training our model, we implemented a custom loss function $\mathcal{L}$ that is calculated by combining three different losses. To compute these three losses and eventually the final loss, we use the ground truth saliency map $Y$, compared not only with the output saliency map $S_{rgb-d}$ of the network, but also with the 4 multi-scale activation maps $A^{m}$ of each of the RGB and the Depth stream, denoted as $(A^{m}_{rgb})$ and $(A^{m}_{d})$ respectively:
\vspace{-0.2cm}
\begin{equation}
\begin{split}
    \mathcal{L} = \mathcal{L}_{sal}(S_{rgb-d}, Y) + (1-\epsilon)\sum_{m = 1}^{4}\mathcal{L}_{rgb}(A^{m}_{rgb}, Y) + \\ (1 - \epsilon)\sum_{m = 1}^{4}\mathcal{L}_{d}(A^{m}_{d}, Y)\hspace{2.5cm}
\end{split}
\end{equation}
where $\epsilon$ is a decaying parameter equal to $\frac{currentEpoch}{\#totalEpochs}$. For the losses $L_{sal}, L_{rgb}$ and $L_{d}$, we calculate three different metrics. We first calculate the cross entropy loss between the generated maps $M$, where $M = S_{rgb-d}, A^{m}_{rgb}, A^{m}_{d}, m = 1,...,4$ and the continuous ground truth saliency map $Y_{c}$ that is obtained by a convolving the binary fixation map $Y_{b}$ of the eye-tracking data with a gaussian kernel:
\begin{equation}
\begin{split}
    \mathcal{L}_{CE}(M, Y_{c}) = -\sum_{x,y}{Y_{c}(x,y)} \odot \log{M(x,y)} \\
    + (1 - Y_{c}(x,y)) \odot (1 - \log{M(x,y)})
\end{split}
\end{equation}
The second metric calculated is the linear Correlation Coefficient (CC) between the map $M$ and the continuous ground truth saliency map $Y_{c}$. The CC metric treats the predicted and the ground truth maps as random variables and uses their covariance $cov$ and standard deviation $\rho$ to calculate their correlation:
\vspace{-0.2cm}
\begin{equation}
    \mathcal{L}_{CC}(M, Y_{c}) = - \frac{cov(M(x,y), Y_{c}(x,y))}{\rho(M(x,y)) \cdot \rho(Y_{c}(x,y))}
\end{equation}
The last metric calculated for the saliency loss, is the Normalized Scanpath Saliency (NSS) metric between the map $M$ and the binary fixation map $Y_{b}$:
\vspace{-0.2cm}
\begin{equation}
    \mathcal{L}_{NSS}(M, Y_{b}) =  -\frac{1}{N_{b}} \sum_{x, y}\tilde{M}(x,y) \odot Y_{b}(x, y)
\end{equation}\\[-0.2cm]
where $\tilde{M}(x, y) = \frac{M(x,y) - \bar{M}(x, y)}{\rho(M(x, y))}$, the normalized map $M$ to zero-mean and unit standard deviation and $N_{b} = \sum_{x, y}Y_{b}(x, y)$, the total number of discrete fixation points in the binary ground truth saliency map. The three initial losses can be written as:
\begin{equation}
\begin{split}
    \mathcal{L}_{sal}(S_{rgb-d}, Y) = w_{1}\mathcal{L}_{CE}(S_{rgb-d}, Y_{c}) + \\ w_{2}\mathcal{L}_{CC}(S_{rgb-d}, Y_{c}) + w_{3}\mathcal{L}_{NSS}(S_{rgb-d}, Y_{b})
\end{split}
\end{equation}
\begin{equation}
\begin{split}
    \mathcal{L}_{rgb}(A^{m}_{rgb}, Y) = w_{1}\mathcal{L}_{CE}(A^{m}_{rgb}, Y_{c}) + \\ w_{2}\mathcal{L}_{CC}(A^{m}_{rgb}, Y_{c}) + w_{3}\mathcal{L}_{NSS}(A^{m}_{rgb}, Y_{b})
\end{split}
\end{equation}
\begin{equation}
\begin{split}
    \mathcal{L}_{d}(A^{m}_{d}, Y) = w_{1}\mathcal{L}_{CE}(A^{m}_{d}, Y_{c}) + \\ w_{2}\mathcal{L}_{CC}(A^{m}_{d}, Y_{c}) + w_{3}\mathcal{L}_{NSS}(A^{m}_{d}, Y_{b})
\end{split}
\end{equation}
where $w_{1}, w_{2}, w_{3}$ are the weights used to get the weighted sum of the three losses.

\subsection{Implementation}

\begin{figure}[t]
\begin{center}
\includegraphics[width=0.5\linewidth]{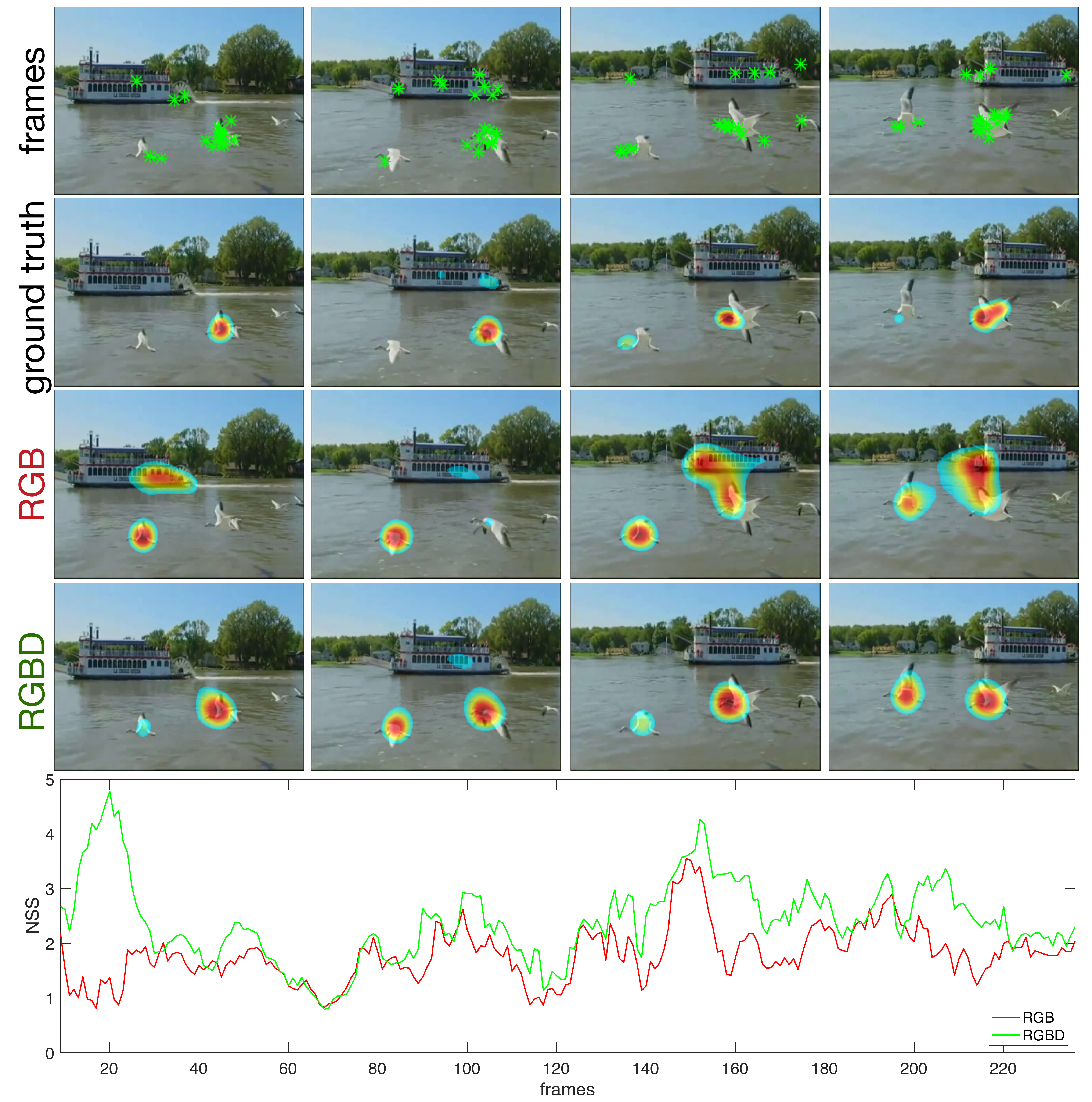}
\end{center}
\vspace{-0.4cm}
   \caption{Sample frames from Coutrot1 database with their eye-tracking data, the corresponding ground truth, RGB-only, and RGB-D saliency maps as produced by ViDaS (RGB and RGB-D). Also NSS curve over time for the two approaches.}
\label{fig:coutrot1}
\vspace{-0.3cm}
\end{figure}

Our implementation and experimentation with the visual network uses as backbone the 3D ResNet-50 architecture \cite{hara2018can} that has showed competitive performance against other deeper architectures for action recognition tasks, in terms of performance and computational budget. As starting point for the trainable parameters $\mathbf{W}_{rgb}, \mathbf{W}_{d}$ of the RGB and the Depth stream respectively, we used the weights from the pretrained model in the Kinetics 400 database.
\noindent \textbf{Training:} For training we employ stochastic gradient descent with momentum 0.9, while we assign a weight decay of 1e-5 for regularization. We have also employed effective batch sizes of 128 samples, and multi-step learning rate. The layers of the DSAM and the Decoder modules are trained using an initial learning rate of 0.0001 while the backbone streams are trained using an initial learning rate of 0.001. The network is trained for 60 epochs. The input data is spatially resized to 112x112 and a sliding window of 16 frames is applied, with the final prediction of the network corresponding to the medium frame. For data augmentation we use random horizontal flipping to the input frames, depthmaps and corresponding ground truth saliency maps with a probability P = 0.5. No other transformations are performed. The weights $w_1, w_2, w_3$ for the saliency loss are selected as 0.1, 2, 1 respectively, after experimentation.



\vspace{-0.2cm}
\section{Experiments}
\vspace{-0.2cm}
\subsection{Datasets}

For training and evaluation of the proposed saliency network, 9 different datasets are employed: DHF1K, Hollywood2, UCF-Sports, DIEM, AVAD, Coutrot1, Coutrot2, SumMe, and ETMD. These databases consist of various types of videos, ranging from very structured small videos to completely unstructured, user-made Youtube videos.
A short description for each database follows.

\noindent\textbf{DHF1K:} DHF1K~\cite{wang2018revisiting} is a large dataset with high content diversity and variable length (from 400 frames to 1200 frames). It includes 1000
videos, out of which 700 are publicly annotated, and 300 are withheld for testing purposes. 

\noindent\textbf{Hollywood-2:} Hollywood2~\cite{marszalek2009actions} contains a collection of short clips with actions performed in Hollywood movies. The dataset was initially developed for human action recognition tasks, but was later adopted for video saliency tasks, after eye-tracking data were collected~\cite{mathe2014actions}. Train and test sets consist of 3100 and 3559 different, non overlapping clips respectively, each one viewed by 12 persons.

\noindent\textbf{UCF-Sports:} UCF-Sports~\cite{soomro2014action, rodriguez2008action} 
similarly to Hollywood2, also contains short clips initially collected for action recognition. Later, eyetracking data from 19 viewers have been recorded~\cite{mathe2014actions}. UCF-sports has been split in train and test set of 104 and 48 non overlapping clips respectively. Both Hollywood2 and UCF-Sports contain shots much shorter and smaller than a DHF1K video sample, ranging from 40 frames to just a single frame per shot.

\noindent\textbf{AVAD:}
AVAD database~\cite{min2017fixation} contains 45 short clips of 5-10 sec duration with several audiovisual scenes, e.g. dancing, guitar playing, bird signing, etc. 
Eye-tracking data from 16 participants have been recorded.

\noindent\textbf{Coutrot databases:}
Coutrot databases~\cite{coutrot_how_2014,Coutrot2016} are split in Coutrot1 and Coutrot2: Coutrot1 contains 60 clips with dynamic natural scenes split in 4 visual categories: one/several moving objects, landscapes, and faces. Eye-tracking data from 72 participants have been recorded. Coutrot2 contains 15 clips of 4 persons in a meeting and the corresponding eye-tracking data from 40 persons.

\noindent\textbf{DIEM:}
DIEM database~\cite{mital+2011} consists of $84$ movies of all sorts, sourced from
publicly  accessible  repositories, including
advertisements, documentaries, game trailers, movie trailers, music videos, news clips, and time-lapse footage. Eye movement data from $42$ participants were recorded. 

\begin{table}[t]
 \centering{
 \resizebox{0.48\textwidth}{!}{
 \begin{tabular}{ | l | c c c c c | }
 \hline
 \multirow{2}{*}{\backslashbox{ \kern0.5em Method \kern-0.1em}{\kern-1.9em Dataset \kern-0.5em}} &  \multicolumn{5}{c|}{Overall}  \\ \cline{2-6} 
 & CC $\uparrow$ & NSS $\uparrow$ & AUC-J $\uparrow$ & sAUC $\uparrow$ & SIM  $\uparrow$ \\ \hline \hline
RGB16SC & 0.5458 & 2.7523 & 0.9025 & 0.6478 & 0.4296  \\ 
Depth16SC & 0.4875 & 2.3494	& 0.89015 & 0.6196 & 0.3890 \\ 
RGBD16SC & 0.5473 & 2.7041 & 0.9022 & 0.6485 & 0.4313\\ \hline
RGB16MF & 0.5859 & 3.0548 & 0.9121 & 0.6584 & 0.4623 \\ \hline
RGB64MF & 0.5950 & 3.1665 & 0.9149 & 0.6624 & 0.4691 \\ 
Depth64MF & 0.5063 & 2.0174 & 0.8954 & 0.6276 & 0.4107 \\ 
RGBD64MF\_ADD & 0.6002 & 3.2074 & 0.9157 & 0.6643 & 0.4689 \\  
RGBD64MF\_CON & 0.5956 & 3.1333 & 0.9152 & 0.6636 & 0.4674 \\ \hline
\textbf{RGBD64MF\_CLL\_MID} & \textbf{0.6041} & \textbf{3.2253} & \textbf{0.9166} & \textbf{0.6654} & \textbf{0.4716} \\ \hline
RGBD64MF\_CLL\_MEG & 0.5968 & 3.1776 & 0.9156 & 0.6640 & 0.4667 \\
RGBD64MF\_CLL\_DIL & 0.5944 & 3.1798 & 0.9147 & 0.6626 & 0.4672 \\ \hline
\end{tabular}
}}
\vspace{0.1cm}
\caption{Ablation study: Different fusion schemes, feature size and depth extraction methods are investigated.}
\vspace{-0.8cm}
\label{table:sal_eval_ablation}
\end{table}

\noindent\textbf{SumMe:}
SumMe database~\cite{gygli+2014, TSIAMI2019} contains $25$ unstructured videos, i.e. mostly user-made videos, from public sources.
Audiovisual eye-tracking data have been collected~\cite{TSIAMI2019} from $10$ viewers.

\noindent\textbf{ETMD:}
ETMD database~\cite{koutras2015,TSIAMI2019} contains $12$ videos from six different hollywood movies. Audiovisual eye-tracking data have been collected~\cite{TSIAMI2019} from $10$ viewers, recorded via an Eyelink eye-tracker.

\vspace{-0.2cm}
\subsection{Experimental Results}

Training has been performed in several different setups by combining data from one or more datasets: For DHF1K, Hollywood2, UCF-Sports and DIEM, the standard splits from literature have been employed~\cite{borji+13,wang2018revisiting,mathe2014actions}. For the other 5 databases, where there is no particular split, the approach adopted in~\cite{Tsiami_2020_CVPR} has been followed (3 non overlapping splits per database and average among splits).

For the evaluation of ViDaS network, we perform an ablation study in order to assess the importance of several parameters and fusion modules, and the contribution of depth in comparison to RGB. Additionally, we compare 3 different depth extraction methods. We then pick our best RGB-D method and compare it to 11 state-of-the-art visual saliency methods (using their publicly available codes and models, re-evaluated), in all 9 databases on the same test data. We also test our RGB-only variant. For all state-of-the-art models, we have employed their best model and re-evaluated the model on the test splits of all databases. During evaluation, we assess ViDaS performance on the various datasets for several training setups. 
Five widely-used saliency evaluation metrics are employed \cite{bylinskii2016different}: CC, NSS, AUC-Judd (AUC-J), shuffled AUC (sAUC) and SIM (similarity). For sAUC we select the negative samples from the union of all viewers' fixations across all other frames in the current video except for the currently processed frame.

\noindent~\textbf{Ablation study:}
Regarding Table~\ref{table:sal_eval_ablation}, ablation study employs 6 datasets: DIEM, AVAD, Coutrot1 and 2, SumMe and ETMD, which are relatively small but with diverse content. Depth extraction is part of the ablation study, and except for the last two rows of Table~\ref{table:sal_eval_ablation}, in all other variants depth extraction has been conducted using MIDAS~\cite{DBLP:journals/corr/abs-1907-01341}.

\noindent\textbf{Methods:} Number 16 or 64 refers to the number of feature maps $S^m$ coming from the DSAM modules. ``SC" refers to Simple Concatenation of the DSAM output maps $S^{m}$ without a decoder, following the approach of~\cite{Tsiami_2020_CVPR}. ``MF" refers to multiscale fusion, i.e. the integration of the Decoder. 

\noindent\textbf{RGB-Depth Fusion:}The ``ADD", ``CON", ``CLL" refer to the different fusion schemes between RGB and Depth. For ``ADD" and ``CON" a simple addition and concatenation respectively of RGB and Depth maps is performed at every scale, and a single common Decoder module was employed for both streams. For ``CLL", each stream is processed individually using its own Decoder module and without any interaction in the different spatio-temporal scales, except for the last layer, where the outputs of each Decoder are concatenated and fused through two convolutional layers. 

\noindent\textbf{Depth extraction:}  Depth is extracted using 3 different depth extraction methods for comparisons and for assessing if a detailed or coarse depth estimation is closer to the concept of saliency. ``MID", ``MEG" and ``DIL" refer to the 3 different depth extraction methods that were investigated, MIDAS~\cite{DBLP:journals/corr/abs-1907-01341}, Megadepth~\cite{li2018megadepth}, and DILATED~\cite{hao2018detail}. 

Our ablation results indicate that multiscale  fusion increased the performance for the RGB-only model. The choice of 64 feature maps along with multiscale fusion further boosted the performance of the RGB model, as well as Depth only model, which in all cases performs worse than the RGB-only model, indicating that Depth can be employed as an additional modality for saliency estimation, but it could not perform equally well on its own. The experimentation with the several fusion schemes indicates that concatenation on the last layer (i.e. late fusion) is more effective and leads to the best performance of our model, probably because each stream learns independently the most it can learn, and the two are combined in the end to produce a single attention map. Lastly, among the three different depth extraction methods, MIDAS leads to our best results. MIDAS produced the most detailed, fine depth estimations among the three, indicating that saliency might be sensitive to depth information. 
For the rest of the paper, by ViDaS [STD] we refer to RGBD64MF\_CLL\_MID,
and by ViDaS [ST] to our RGB-only variant. 
An example comparison of the two versions is depicted in Fig.~\ref{fig:coutrot1} along with the original frames and the ground truth saliency maps. Also the NSS curve is depicted over time. The particular frame has many levels of depth and indeed the RGB-D version captures saliency better than the RGB-only version.

\begin{table*}[t!]
 \centering{
 \resizebox{0.98\textwidth}{!}{
 \begin{tabular}{ l | c c c c c || c c c c c || c c c c c}
 \hline
 \multirow{2}{*}{\backslashbox{ \kern0.5em Method \kern-0.5em}{\kern-1.9em Dataset \kern-0.5em}} &  \multicolumn{5}{c||}{DHF1K}   &  \multicolumn{5}{c||}{Hollywood-2} & \multicolumn{5}{c}{UCFsports} \\ \cline{2-16} 
 & CC $\uparrow$ & NSS $\uparrow$ & AUC-J $\uparrow$ & sAUC $\uparrow$ & SIM  $\uparrow$ & CC $\uparrow$ & NSS $\uparrow$ & AUC-J $\uparrow$ & sAUC $\uparrow$ & SIM  $\uparrow$ & CC $\uparrow$ & NSS $\uparrow$ & AUC-J $\uparrow$ & sAUC $\uparrow$ & SIM  $\uparrow$\\ \hline \hline
ViDaS [STD] tDHF1K & 0.4891 & 2.75 & 0.9071 & 0.6924 & 0.3790 &
 0.5982 & 2.82 & 0.9128 & 0.5396 & 0.4903 &
 0.5673 &  2.73 & 0.8981 & 0.5972 & 0.4647\\
ViDaS [ST] tDHF1K & 0.4864 & 2.73 & 0.9079 & 0.6913 & 0.3787 &
 0.5901 & 2.78 & 0.9106 & 0.5381 & 0.4887 &
 0.5538 & 2.66 & 0.8985 & 0.5963 & 0.4565 \\
ViDaS [STD] tHOLLY & 0.4366 & 2.42 & 0.8913 & 0.6728 & 0.3399 &
0.6457 & 2.96 & 0.9173 & 0.5409 & 0.5262 &
0.5462 & 2.57 & 0.8801 & 0.5959 & 0.4549\\
ViDaS [ST] tHOLLY & 0.4338 & 2.40 & 0.8926 & 0.6702 & 0.3420 &
 0.6425 & 2.94 &  0.9168 & 0.5402 & 0.5255 &
 0.5400 & 2.53 & 0.8844 & 0.5915 & 0.4537\\
ViDaS [STD] tUCF & 0.4074 & 2.27 & 0.8815 & 0.6712 & 0.3109 &
0.4747 & 2.06 & 0.8728 & 0.5219 & 0.3781 &
0.6360 & 3.26 & \textbf{0.9160} & 0.6477 & 0.5241\\
ViDaS [ST] tUCF & 0.3928 & 2.20 & 0.8764 & 0.6622 & 0.3129 &
0.4471 & 1.92 & 0.8609 & 0.5188 & 0.3725 &
 0.6317 & 3.23 & 0.9124 & 0.6467 & 0.5242\\
ViDaS [STD] tUHD & 0.4778 & 2.69 & 0.9058 & 0.6876 & 0.3724 &
0.6462 & 3.00 & 0.9184 & 0.5427 & 0.5283 &
\textbf{0.6463} & 3.26 & 0.9111 & 0.6383 & \textbf{0.5331}\\
ViDaS [ST] tUHD & 0.4798  & 2.70 & 0.9056 & 0.6869 & 0.3743 &
 \textbf{0.6653} & 2.90 & 0.9146 & 0.5297 & \textbf{0.5397} &
 0.6272 & 3.12 & 0.9089 & 0.6310 & 0.5194\\
ViDaS [STD] tSTAViS & 0.4736 & 2.64 & 0.9047 & 0.6924 & 0.3583 &
 0.6202 & 2.77 & 0.9131 & 0.5324 & 0.5003 &
0.5818 & 2.78 & 0.9020 & 0.6189 & 0.4664\\ 
ViDaS [ST] tSTAViS & 0.4693 & 2.61 & 0.9034 &  0.6884 & 0.3582 &
0.6141 & 2.75 & 0.9123 & 0.5312 & 0.4993 &
0.5762 & 2.72 & 0.8954 & 0.6082 & 0.4662\\ \hline
DeepNet \cite{Pan_2016_CVPR} [S]& 0.2969 & 1.58 & 0.8421 & 0.6432 & 0.1878 &
0.4163 & 1.89 & 0.8717 & 0.5416 & 0.2851 &
0.4121 & 1.89 & 0.8609 & 0.6162 & 0.2844\\
DVA \cite{wang2018deep} [S] & 0.3592 & 2.06 & 0.8609 & 0.6572 & 0.2462 &
0.4644 & 2.44 & 0.8806 & 0.5529 & 0.3500 &
0.4495 & 2.37 & 0.8706 & 0.6207 & 0.3288\\
SAM \cite{cornia2018predicting} [S] & 0.3684 & 2.12 & 0.8680 & 0.6562 & 0.2918 &
0.4798 & 2.61 & 0.8858 & 0.5552 & 0.4009 &
0.4941 & 2.75 & 0.8854 & 0.6272 & 0.4036\\ 
SalGAN \cite{Pan_2017_SalGAN} [S] & 0.3533 & 1.95 & 0.8626 & 0.6732 &  0.2515 & 
0.4534 & 2.19 & 0.8761 & 0.5540 &  0.3475 &
0.4388 & 2.10 & 0.8674 & 0.6240 & 0.3254\\
ACLNet \cite{wang2018revisiting,wang2019revisiting} [ST] & 0.4167 & 2.30 & 0.8883  & 0.6523  & 0.3008 &
0.5954 & 3.06 & 0.9179 & 0.5428 & 0.4855 &
0.5070  & 2.54 & 0.8977 & 0.5908 & 0.4058  \\
DeepVS \cite{jiang2018deepvs} [ST] & 0.3500 & 1.97 & 0.8561 & 0.6405 & 0.2622 &
 0.4769 & 2.48 & 0.8883 & 0.5481 & 0.3857 &
 0.4550 & 2.31 & 0.8703 & 0.6136 & 0.3682\\ 
TASED \cite{Min_2019_ICCV} [ST] &  \textbf{0.5142} & \textbf{2.87} & \textbf{0.9130} & \textbf{0.7123} & 0.3592 &
 0.5622 & 2.77 & 0.9138 & 0.5397 & 0.4372 &
 0.4943 & 2.31 & 0.8884 & 0.5528 & 0.4027 \\
Unisal \cite{unisal} [ST] & 0.4778 & 2.75 & 0.8994 &  0.6759 & 0.3815 &
0.6158 & \textbf{3.40} & \textbf{0.9217} & \textbf{0.5739} & 0.4961 &
0.6254 & \textbf{3.38} & 0.9117 & \textbf{0.6536} & 0.5104\\
STRANet \cite{lai2019video} [ST] &  0.4617 & 2.58 & 0.8971 & 0.6727 & 0.3568 &
0.6010 & 3.19 & 0.8735 & 0.5520 & 0.4922 &
  0.5635 & 2.85 & 0.9067 & 0.6135 & 0.4639 \\
SALEMA \cite{Linardos2019} [ST] & 0.4939  & 2.86  & 0.9064  & 0.6866 & \textbf{0.3919} & 
 0.5531 & 2.98 & 0.9085 & 0.5545 & 0.4622 & 
 0.5551 & 2.93 & 0.9014 & 0.6218 & 0.4614\\
STAViS [ST]~\cite{Tsiami_2020_CVPR} & 0.4312 & 2.35 & 0.8936 & 0.6789 & 0.3139 &
0.5898 & 2.59 & 0.9085 & 0.5303 & 0.4684 &
0.5376 & 2.44 & 0.8906 & 0.6051 & 0.4239 \\ \hline 
\end{tabular}
}}
\vspace{0.1cm}
\caption{Evaluation results for saliency in DHF1K validation set, Hollywood2 and UCFsports databases. The proposed method's (ViDaS [STD] and the RGB-only variant  [ST]) results are depicted for different training setups. [STD] stands for spatio-temporal plus depth, [ST] for spatio-temporal visual models while [S] denotes spatial only models.}
\label{table:sal_eval1}
\vspace{-0.5cm}
\end{table*}

\begin{table*}[t!]
 \centering{
 \resizebox{0.98\textwidth}{!}{
 \begin{tabular}{ l | c c c c c || c c c c c || c c c c c}
 \hline
 \multirow{2}{*}{\backslashbox{ \kern0.5em Method \kern-0.5em}{\kern-1.9em Dataset \kern-0.5em}} &  \multicolumn{5}{c||}{DIEM}   &  \multicolumn{5}{c||}{Coutrot1} & \multicolumn{5}{c}{Coutrot2} \\ \cline{2-16} 
 & CC $\uparrow$ & NSS $\uparrow$ & AUC-J $\uparrow$ & sAUC $\uparrow$ & SIM  $\uparrow$ & CC $\uparrow$ & NSS $\uparrow$ & AUC-J $\uparrow$ & sAUC $\uparrow$ & SIM  $\uparrow$ & CC $\uparrow$ & NSS $\uparrow$ & AUC-J $\uparrow$ & sAUC $\uparrow$ & SIM  $\uparrow$\\ \hline \hline
ViDaS [STD] tDHF1K & 0.5772 & 2.28 & 0.8819 & 0.6540 & 0.4896 &
0.4662 & 2.08 & 0.8657 & 0.5720 & 0.3927 &
0.5331 & 3.77 & 0.9304 & 0.6389 & 0.3921\\
ViDaS [ST] tDHF1K & 0.5609 & 2.22 & 0.8826 & 0.6515 & 0.4839 &
0.4600 & 2.06 & 0.8649 & 0.5740 & 0.3918 &
0.5672 & 3.94 & 0.9250 & 0.6569 & 0.3881 \\
ViDaS [STD] tHOLLY & 0.5640 & 2.23 & 0.8794 & 0.6504 & 0.4763 &
0.4621 & 2.08 & 0.8594 & 0.5704 & 0.3927 &
0.4910 & 3.31 & 0.9224 & 0.6464 & 0.3652\\
ViDaS [ST] tHOLLY & 0.5585 & 2.20 & 0.8766 & 0.6459 & 0.4761 & 
0.4526 & 2.03 & 0.8565 & 0.5650 & 0.3872 &
0.4742 & 3.19 & 0.9137 & 0.6451 & 0.3420\\
ViDaS [STD] tUCF & 0.4732 & 1.88 & 0.8506 & 0.6386 & 0.4071 &
0.3844 & 1.69 & 0.8403 & 0.5656 & 0.3358 &
0.4959 & 3.39 & 0.9295 & 0.6795 & 0.3190\\
ViDaS [ST] tUCF & 0.4651 & 1.83 & 0.8561 & 0.6357 &  0.4067 &
0.3730 & 1.68 & 0.8331 & 0.5737 & 0.3358&
0.4566 & 3.02 & 0.9043 & 0.6791 & 0.2886\\
ViDaS [STD] tUHD & 0.5693 & 2.26 & 0.8834 & 0.6521 & 0.4829 &
0.4791 & 2.18 & 0.8635 & 0.5759 & 0.4035 & 
0.4258 & 3.01 & 0.9250 & 0.6329 & 0.3437\\
ViDaS [ST] tUHD & 0.5721 & 2.27 & 0.8834 & 0.6551 & 0.4870 &
0.4877 & 2.22 & 0.8705 & 0.5787 & 0.4076 &
0.5295 & 3.73 & 0.9351 & 0.6555 & 0.3894\\
ViDaS [STD] tSTAViS & \textbf{0.6387} & \textbf{2.50} & \textbf{0.8995} & \textbf{0.6848} & \textbf{0.5278} &
\textbf{0.5256} & \textbf{2.37} & \textbf{0.8786} & \textbf{0.5891} & \textbf{0.4300} &
\textbf{0.7701} & \textbf{5.71} & \textbf{0.9633} & 0.7147 & 0.5577\\ 
ViDaS [ST] tSTAViS & 0.6342 & 2.48 & 0.8963 & 0.6834 &  0.5254 &
0.5056 & 2.28 & 0.8754 & 0.5841 &  0.4196 &
0.7679 & 5.64 & 0.9627 & \textbf{0.7150} & \textbf{0.5589} \\ \hline
DeepNet \cite{Pan_2016_CVPR} [S]& 0.4075 & 1.52 & 0.8321 & 0.6227 & 0.3183& 
0.3402 & 1.41 & 0.8248 & 0.5597 & 0.2732 & 
0.3012 & 1.82 & 0.8966 & 0.6000 & 0.2019 \\
DVA \cite{wang2018deep} [S] & 0.4779 & 1.97 & 0.8547 & 0.641 & 0.3785 & 
0.4306 & 2.07 & 0.8531 & 0.5783 & 0.3324
& 0.4634 & 3.45 & 0.9328 & 0.6324 & 0.2742 \\
SAM \cite{cornia2018predicting} [S] & 0.4930 & 2.05 & 0.8592 & 0.6446 & 0.4261 &
0.4329 & 2.11 & 0.8571 & 0.5768 & 0.3672 & 
0.4194 & 3.02 & 0.9320 & 0.6152 & 0.3041 \\ 
SalGAN \cite{Pan_2017_SalGAN} [S] & 0.4868 & 1.89 & 0.8570 & 0.6609 & 0.3931 &
0.4161 & 1.85 & 0.8536 & 0.5799 & 0.3321 & 
0.4398 & 2.96 & 0.9331 & 0.6183 & 0.2909 \\
ACLNet \cite{wang2018revisiting,wang2019revisiting} [ST] & 0.5229 & 2.02 & 0.8690 & 0.6221 & 0.4279 & 
0.4253 & 1.92 & 0.8502 & 0.5429 & 0.3612 & 
0.4485 & 3.16 & 0.9267 & 0.5943 & 0.3229 \\
DeepVS \cite{jiang2018deepvs} [ST] & 0.4523 & 1.86 & 0.8406 & 0.6256 & 0.3923 &
0.3595 & 1.77 & 0.8306 & 0.5617 & 0.3174 & 
0.4494 & 3.79 & 0.9255 & 0.6469 & 0.2590 \\ 
TASED \cite{Min_2019_ICCV} [ST] & 0.5579 & 2.16 & 0.8812 & 0.6579 & 0.4615 &
0.4799 & 2.18 & 0.8676 & 0.5808 & 0.3884 & 
0.4375 & 3.17 & 0.9216 & 0.6118 & 0.3142 \\ 
Unisal \cite{unisal} [ST] & 0.5711 & 2.36 & 0.8789 & 0.6435 & 0.4822 &
0.4248 & 2.07 & 0.8489 & 0.5642 & 0.3714 &
0.3647 & 2.82 & 0.9301 & 0.5986 & 0.3012 \\
SALEMA \cite{Linardos2019} [ST] &  0.5180 & 2.13 & 0.8638 & 0.6320 & 0.4515 &
0.4334 & 2.05 & 0.8505 & 0.5608 & 0.3747 &
0.4671 & 3.67 & 0.9273 & 0.6162 & 0.3402\\
STAViS [ST]~\cite{Tsiami_2020_CVPR} & 0.5665 & 2.19 & 0.8792 & 0.6648 & 0.4719 & 0.4587 & 1.99 & 0.8617 & 0.5764 & 0.3842 &
0.6529 & 4.19 & 0.9405 & 0.6895 & 0.4470 \\ 
STAViS [STA]~\cite{Tsiami_2020_CVPR} & 0.5795 & 2.26 & 0.8838 & 0.6741 & 0.4824 & 
0.4722 & 2.11 & 0.8686 & 0.5847 & 0.3935 &
0.7349 & 5.28 & 0.9581 & 0.7106 & 0.5111 \\  
\hline 
\end{tabular}
}}
\vspace{0.1cm}
\caption{Evaluation results for saliency in DIEM, Coutrot1 and Coutrot2 databases. The proposed method's (ViDaS [STD] and the RGB-only variant  [ST]) results are depicted for different training setups. [STD] stands for spatio-temporal plus depth, [STA] for spatio-temporal plus audio, [ST] for spatio-temporal visual models while [S] denotes spatial only models.}
\label{table:sal_eval2}
\vspace{-1.0cm}
\end{table*}

\begin{figure}[t]
\begin{center}
\includegraphics[width=0.45\linewidth]{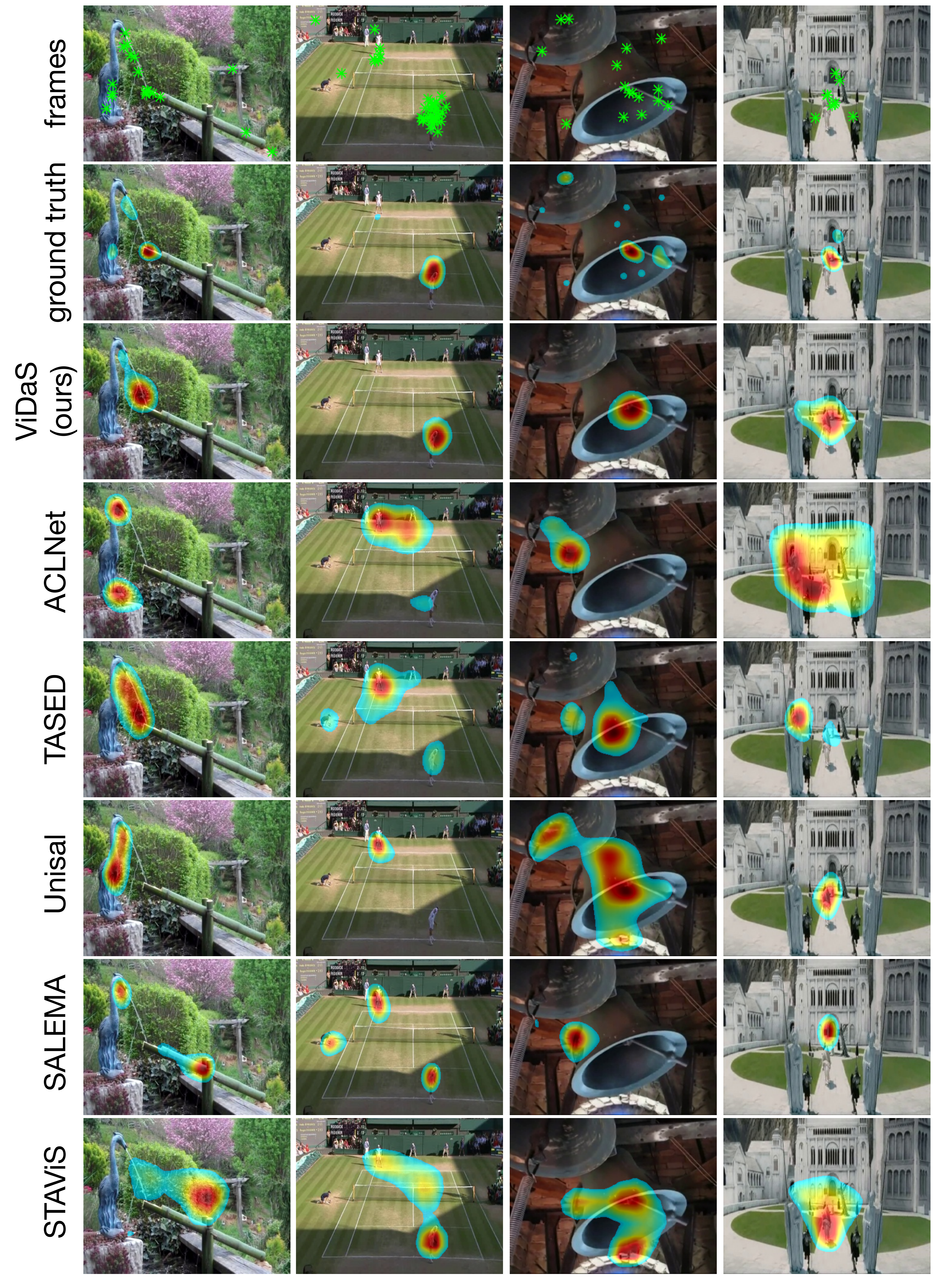}
\end{center}
\vspace{-0.4cm}
   \caption{Sample frames from several databases with their eye-tracking data, the corresponding ground truth, ViDaS RGB-D network and several other state-of-the-art methods.}
\label{fig:sota_fig}
\vspace{-0.4cm}
\end{figure}

\noindent~\textbf{Comparison to state-of-the-art:}
Extensive comparisons with 11 different state-of-the-art saliency methods on 9 different databases are depicted for the five metrics per database, in Tables~\ref{table:sal_eval1},~\ref{table:sal_eval2} and~\ref{table:sal_eval3}. The models were not retrained, since in some cases the code is not available, but the published pretrained models were used to consistently evaluate the results in the same way. Since all of the other methods have been trained on combinations of SALICON, DHF1K, Hollywood-2 and UCF-Sports datasets, we included 5 different training set-ups (on DHF1K only, on Hollywood2, on UCF, on a combination of these three denoted by UHD and on the rest 6 databases, denoted by STAViS as depicted in Tables~\ref{table:sal_eval1},~\ref{table:sal_eval2}, and~\ref{table:sal_eval3}) to enable fair comparisons, as well as the performance assessment of the model in unseen or seen datasets. For each training setup, we have also trained our RGB-only variant. Overall, ViDaS RGB-D network achieves the best or competitive performance on the various datasets. In Table~\ref{table:sal_eval1} UNISAL performs better than ViDaS in some metrics, perhaps because it has been pretrained on saliency datasets (SALICON, DHF1K), that seems to boost performance, whereas ViDaS uses the pretrained weights of Kinetics400 (which is an action recognition dataset) as a starting point. Also some methods have tuned their parameters to these most widely used datasets, whereas our method was trained in a more robust way. TASED method that performs better in DHF1K, employs a 32-frame temporal length, compared to ours which is 16. For visualization purposes, in Fig.~\ref{fig:sota_fig} sample frames are presented with their corresponding eye-tracking data, ground truth saliency maps, and the corresponding saliency maps from our proposed ViDaS RGB-D network and other state-of-the-art methods: ACLNet, TASED, Unisal, SALEMA and STAViS. It can easily be observed that our results are closer to the ground truth, especially when frames have several levels of depth.

\begin{table*}[t!]
 \centering{
 \resizebox{0.98\textwidth}{!}{
 \begin{tabular}{ l | c c c c c || c c c c c || c c c c c}
 \hline
 \multirow{2}{*}{\backslashbox{ \kern0.5em Method \kern-0.5em}{\kern-1.9em Dataset \kern-0.5em}} &  \multicolumn{5}{c||}{AVAD}   &  \multicolumn{5}{c||}{SumMe} & \multicolumn{5}{c}{ETMD} \\ \cline{2-16} 
 & CC $\uparrow$ & NSS $\uparrow$ & AUC-J $\uparrow$ & sAUC $\uparrow$ & SIM  $\uparrow$ & CC $\uparrow$ & NSS $\uparrow$ & AUC-J $\uparrow$ & sAUC $\uparrow$ & SIM  $\uparrow$ & CC $\uparrow$ & NSS $\uparrow$ & AUC-J $\uparrow$ & sAUC $\uparrow$ & SIM  $\uparrow$\\ \hline \hline
ViDaS [STD] tDHF1K & 0.6270 & 3.45 & 0.9204 & 0.5945 & 0.4941 &
0.4352 & 2.18 & 0.8916 & 0.6492 & 0.3602 &
0.5252 & 2.74 & 0.9247 & 0.7000 & 0.4176\\
ViDaS [ST] tDHF1K & 0.6329 & 3.46 & 0.9227 & 0.5951 & \textbf{0.4989} &
0.4417 & 2.20 & 0.8938 & 0.6489 & 0.3655 &
0.5108 & 2.66 & 0.9215 & 0.6921 & 0.4107 \\
ViDaS [STD] tHOLLY & 0.6088 & 3.28 & 0.9191 & 0.5943 & 0.4799 &
0.4092 & 2.01 & 0.8801 & 0.6412 & 0.3436 & 
0.5378 & 2.78 & 0.9253 & 0.7109 & 0.4221\\
ViDaS [ST] tHOLLY & 0.6091 & 3.26 & 0.9176 & 0.5901 & 0.4821 &
0.3942 & 1.92 & 0.8760 & 0.6291 & 0.3367 &
0.5287 & 2.72 & 0.9236 & 0.7040 & 0.4181\\
ViDaS [STD] tUCF & 0.5389 & 2.86 & 0.9124 & 0.5917 & 0.4064 &
0.3601 & 1.84 & 0.8581 & 0.6342 & 0.3004 &
0.4234 & 2.20 & 0.8962 & 0.6900 & 0.3188\\
ViDaS [ST] tUCF & 0.4755 & 2.46 & 0.9031 & 0.5806 & 0.3793 &
0.3417 & 1.75 & 0.8542 & 0.6317 & 0.3014 &
0.4025 & 2.08 & 0.8893 & 0.6848 & 0.3181\\
ViDaS [STD] tUHD & 0.6374 & 3.51 & 0.9241 & 0.5960 & 0.4983 &
0.4249 & 2.12 & 0.8891 & 0.6478 & 0.3550 &
0.5450 & 2.84 & 0.9284 & 0.7146 & 0.4283\\
ViDaS [ST] tUHD & 0.6270 & 3.41 & 0.9216 & 0.5933 &  0.4924 &
0.4240 & 2.11 & 0.8886 & 0.6444 & 0.3552 &
0.5404 & 2.80 & 0.9268 & 0.7128 & 0.4240\\
ViDaS [STD] tSTAViS & \textbf{0.6481} & 3.45 & \textbf{0.9262} & \textbf{0.5991} & 0.4976 &
\textbf{0.4541} & 2.24 & \textbf{0.8973} & 0.6675 & 0.3627 &
\textbf{0.5882} & \textbf{3.07} & \textbf{0.9349} & 0.7374 & \textbf{0.4537}\\ 
ViDaS [ST] tSTAViS & 0.6371 & 3.39 & 0.9242 & 0.5955 & 0.4988 &
0.4458 & 2.20 & 0.8971 & 0.6645 & 0.3617 &
0.5791 & 3.02 & 0.9336 & 0.7321 & 0.4499 \\ \hline
DeepNet \cite{Pan_2016_CVPR} [S]& 0.3831 & 1.85 & 0.8690 & 0.5616 & 0.2564& 
0.3320 & 1.55 & 0.8488 & 0.6451 & 0.2274 & 
0.3879 & 1.90 & 0.8897 & 0.6992 & 0.2253 \\
DVA \cite{wang2018deep} [S] & 0.5247 & 3.00 & 0.8887 & 0.5820 & 0.3633 & 
0.3983 & 2.14 & 0.8681 & 0.6686 & 0.2811
& 0.4965 & 2.72 & 0.9039 & 0.7288 & 0.3165 \\
SAM \cite{cornia2018predicting} [S] & 0.5279 & 2.99 & 0.9025 & 0.5777 & 0.4244 &
0.4041 & 2.21 & 0.8717 & 0.6728 & 0.3272 & 
0.5068 & 2.78 & 0.9073 & 0.7310 & 0.3790 \\ 
SalGAN \cite{Pan_2017_SalGAN} [S] & 0.4912 & 2.55 & 0.8865 & 0.5799 & 0.3608 &
0.3978 & 1.97 & 0.8754 & \textbf{0.6882} & 0.2897 & 
0.4765 & 2.46 & 0.9035 & \textbf{0.7463} & 0.3117 \\ 
ACLNet \cite{wang2018revisiting,wang2019revisiting} [ST] & 0.5809 & 3.17 & 0.9053 & 0.5600 & 0.4463 & 
0.3795 & 1.79 & 0.8687 & 0.6092 & 0.2965 & 
0.4771 & 2.36 & 0.9152 & 0.6752 & 0.3290 \\
DeepVS \cite{jiang2018deepvs} [ST] & 0.5281 & 3.01 & 0.8968 & 0.5858 & 0.3914 &
0.3172 & 1.62 & 0.8422 & 0.6120 & 0.2622 & 
0.4616 & 2.48 & 0.9041 & 0.6861 & 0.3495 \\ 
TASED \cite{Min_2019_ICCV} [ST] & 0.6006 & 3.16 & 0.9146 & 0.5898 & 0.4395 &
0.4288 & 2.10 & 0.8840 & 0.6570 & 0.3337 & 
0.5093 & 2.63 & 0.9164 & 0.7117 & 0.3660 \\ 
Unisal \cite{unisal} [ST] & 0.6220 & \textbf{3.69} & 0.9143 & 0.5924 & 0.4969 &
0.4459 & \textbf{2.37} & 0.8899 & 0.6480 & \textbf{0.3725} &
0.5432 & 2.96 & 0.9275 & 0.7093 & 0.4287\\
SALEMA \cite{Linardos2019} [ST] &  0.5500 & 3.17 & 0.9067 & 0.5738 & 0.4441 &
0.4073 & 2.10 & 0.8772 & 0.6290 & 0.3440 &
0.5108 & 2.76 & 0.9192 & 0.6955 & 0.4057\\
STAViS [ST]~\cite{Tsiami_2020_CVPR} & 0.6041 & 3.07 & 0.9157 & 0.5900 & 0.4431 & 0.4180 & 1.98 & 0.8848 & 0.6477 & 0.3325 &
0.5602 & 2.84 & 0.9290 & 0.7278 & 0.4121 \\ 
STAViS [STA]~\cite{Tsiami_2020_CVPR} & 0.6086 & 3.18 & 0.9196 & 0.5936 & 0.4578 &
0.4220 & 2.04 & 0.8883 & 0.6562 & 0.3373 &
0.5690 & 2.94 & 0.9316 & 0.7317 & 0.4251 \\ \hline 
\end{tabular}
}}
\vspace{0.1cm}
\caption{Evaluation results for saliency in AVAD, SumMe and ETMD databases. The proposed method's (ViDaS [STD] and the RGB-only variant  [ST]) results are depicted for different training setups. [STD] stands for spatio-temporal plus depth, [STA] for spatio-temporal plus audio, [ST] for spatio-temporal visual models while [S] denotes spatial only models.}
\vspace{-0.9cm}
\label{table:sal_eval3}
\end{table*}

\subsection{Discussion}
Results in the various datasets indicate that depth as a complimentary modality for saliency estimation indeed boosts performance, especially when frames contain many levels of depth, e.g. Coutrot1 and UCF sports. In almost all cases, the RGB-D ViDaS version performs consistently better than the RGB-only version, endowing the model with robustness and smoother estimations across time. An interesting finding is that ViDaS model performance is not degraded even on unseen datasets, and in such cases the existence of depth makes a even bigger difference. For example, in AVAD, SumMe, Coutrot2 datasets, the best performance is achieved by training on these datasets, but the model trained on DHF1K or UHD still achieves a competitive performance, whereas other methods like UNISAL, SALEMA and TASED that exhibited good performance in seen datasets (e.g. DHF1K), when tested in unseen data (e.g. Coutrot2), their performance is not consistently good. To sum up, results indicate that ViDaS RGB-D network can well generalize into unseen datasets, without a large compromise in performance, confirming its potential for modeling saliency ``in-the-wild".

\vspace{-0.4cm}
\section{Conclusions}
\vspace{-0.2cm}

We presented ViDaS, a novel video depth-aware saliency network that efficiently predicts fixations in videos, by combining an RGB and a Depth stream in order to produce a single saliency map. Network performance has been extensively evaluated in various datasets with highly diverse content. Results for 5 different metrics in 9 different databases and comparison with 11 state-of-the-art methods indicate that depth can endow an RGB network with robustness and performance boost. Our RGB-D method achieves the best or competitive performance in all cases. Also, its better performance in unseen datasets indicate its appropriateness for estimating saliency ``in-the-wild".

\clearpage
%
%
\bibliographystyle{splncs04}
\bibliography{eccv2022submission}
\end{document}